\documentclass[11pt]{article}
\usepackage{lineno}

\usepackage{geometry}
\usepackage{amssymb}
\usepackage{natbib}
\renewcommand\cite{\citep}
\usepackage{url}

\usepackage{times}
\usepackage{latexsym}

\usepackage{amsmath}

\usepackage[T1]{fontenc}

\usepackage[utf8]{inputenc}

\usepackage{microtype}

\usepackage{inconsolata}

\usepackage{graphicx}

\usepackage{booktabs}
\usepackage[framemethod=TikZ]{mdframed}
\usepackage{tikz}
\usetikzlibrary{shapes.geometric, positioning, arrows.meta, fit, backgrounds, calc}
\usepackage{xcolor}
\usepackage{amsmath}
\AtBeginDocument{%
  \setlength{\abovedisplayskip}{6pt plus 2pt minus 4pt}%
  \setlength{\abovedisplayshortskip}{0pt plus 2pt}%
  \setlength{\belowdisplayskip}{6pt plus 2pt minus 4pt}%
  \setlength{\belowdisplayshortskip}{4pt plus 2pt minus 2pt}%
}
\usepackage{algpseudocode}
\usepackage{algorithm}
\usepackage{tcolorbox}
\tcbuselibrary{skins}
\usepackage{multirow}
\tcbset{
  colback=gray!5,
  colframe=gray!50,
  coltitle=black,
  boxrule=0.4pt,
  arc=2pt,
}
\usepackage{mdframed}
\usepackage{float}
\usepackage{enumitem}

\definecolor{dullgreen}{RGB}{60,120,60}
\usepackage[colorlinks=true,linkcolor=blue,citecolor=blue,urlcolor=blue]{hyperref}

\newcommand{\name}{\textsc{NN-PPI}}

\newcommand{\claudeopus}{Claude Opus 4.6}
\newcommand{\gemmasmall}{Gemma~3 270M}
\newcommand{\xlmrobertalft}{XLM-RoBERTA-Large (FT)}
\newcommand{\gemmaoneb}{Gemma~3 1B}
\newcommand{\gemmafourb}{Gemma~3 4B}
\newcommand{\gptmodel}{GPT-5.2}

\newcommand{\mbaseline}{Baseline}
\newcommand{\mknn}{KNN}
\newcommand{\mknnppi}{\name{}}

\title{Calibrating Small Language Models for Claim Check-Worthiness Detection}

\author{
  \textbf{Pratuat Amatya}\textsuperscript{1,2} \quad
  \textbf{Venktesh V}\textsuperscript{3} \quad
  \textbf{Vinay Setty}\textsuperscript{1,2}
  \\[4pt]
  \small{
    \href{mailto:pratuat@factiverse.ai}{\textcolor{black}{\texttt{pratuat@factiverse.ai}}} \quad
    \href{mailto:venktesh.viswanathan@dsv.su.se}{\textcolor{black}{\texttt{venktesh.viswanathan@dsv.su.se}}} \quad
    \href{mailto:vinay@factiverse.ai}{\textcolor{black}{\texttt{vinay@factiverse.ai}}}
  }
  \\
  \\[4pt]
  \textsuperscript{1}Factiverse AS \quad
  \textsuperscript{2}University of Stavanger \quad
  \textsuperscript{3}Stockholm University
}

\date{}

\begin{document}
\maketitle
\begin{abstract}
Assessing claim check-worthiness is an essential first step in automated fact-checking pipelines. This work is motivated by a real deployment challenge at an early-stage startup: running large language models (LLMs) over every incoming claim is cost- and latency-prohibitive, yet smaller models sacrifice accuracy. We propose \name{}, a pointwise extension of Prediction-Powered Inference (PPI) that calibrates model predictions at inference time as a lightweight post-hoc layer, without re-training the underlying model. \name{} achieves weighted F1 gains ranging from \textbf{12\%} to \textbf{33.80\%} depending on the size and performance of the baseline model, bringing SLMs on par with larger LLMs. Beyond few-shot SLMs, \name{} further improves a production-deployed fine-tuned model, demonstrating that residual calibration is complementary to supervised fine-tuning. By recovering LLM-level accuracy from models that are an order of magnitude cheaper to serve, it makes accurate check-worthiness detection substantially cheaper to operate at scale. Our code and data can be found at \url{https://anonymous.4open.science/r/arr-claim-worthiness-F237/}
\end{abstract}

\section{Introduction}
\label{sec:introduction}

The rapid spread in misinformation in the digital age has been identified as one of the critical issues by World Economic Forum \cite{world_economic_forum}.
Automated fact-checking advances have been made in recent years to combat the surge in misinformation.

This work is motivated by the practical demands of operating a commercial fact-checking service at an early-stage startup. Claim check-worthiness is the first filtering stage: it scans high-velocity streams of news articles, political debates, and social media posts to surface claims worth routing to human fact-checkers. Two constraints dominate this setting. \emph{First, cost and latency}: invoking large LLMs on every incoming claim is prohibitive at production volumes, pushing practitioners toward small, cheap-to-serve models that lag in accuracy. \emph{Second, adaptivity}: editorial notions of check-worthiness evolve over time and across topics, so a deployed system must adapt without costly re-training. This raises the central question: can we retain the low serving cost of small models while recovering the accuracy of large LLMs?

Existing works on claim worthiness detection are primarily focused on fine-tuning pre-trained models on well-curated claims or focus on extracting claims on social media and news articles \cite{stammbach-etal-2023-environmental,sheikhi-etal-2023-automated}. While, they require a considerable amount of manually annotated training samples, more recently, Large Language Models (LLMs) have been employed in a few-shot or zero-shot setting to aid in claim worthiness detection \cite{hyben2024multilingualmultitopicalbenchmarkfinetuned,majer-snajder-2024-claim,ni-etal-2024-afacta}. Fact-checking organizations also frequently update the claimworthiness guidelines, rendering it easy for adaptation using few-shot or zero-shot methods \cite{majer-snajder-2024-claim}. 

However, Large Language models are unreliable \cite{si-etal-2024-large}, sensitive to prompt variations \cite{zhuo-etal-2024-prosa}. Hence, their predicted outputs being uncalibrated and align poorly with human judgments and notions of what constitutes checkworthy claims \cite{majer-snajder-2024-claim,hyben2024multilingualmultitopicalbenchmarkfinetuned}. 

Hence, our work focuses on calibration of LLM predicted outputs to align them with human judgments using a small manually annotated calibration set. Inspired by Prediction-Powered Inference \cite{angelopoulos2023predictionpoweredinference,angelopoulos2024ppiefficientpredictionpoweredinference}, which is usually only employed for calibrating population or system-level metrics, we propose Nearest Neighbor Prediction-Powered Inference (\name{}): a pointwise extension that provides per-instance confidence intervals, requiring a new method for computing residuals and variance that is not a trivial extension of PPI. We further show that \name{} improves upon a production-deployed fine-tuned XLM-RoBERTa-Large model, demonstrating that residual calibration is complementary to supervised fine-tuning.
We address the research questions: 

\textbf{RQ 1}: How does \name{} calibration affect check-worthiness prediction performance across model scales?

\textbf{RQ 2}: How does \name{} perform relative to an uncalibrated baseline and a plain KNN baseline across datasets?

\textbf{RQ 3}: How does neighbor size $k$ affect \name{} calibration performance?

\section{Related Works}

 The Claim Checkworthiness detection is the first stage of an automated fact-checking pipeline which entails checking which parts of the input are deemed necessary for fact-checking \cite{majer-snajder-2024-claim}. The checkworthiness detection task has usually been framed as a classification task with existing works adopting classical supervised machine learning approaches \cite{supervised_CW,wright-augenstein-2020-claim,gencheva-etal-2017-context}. Alternate formulations of the claim worthiness detection task include claim ranking \cite{jaradat-etal-2018-claimrank,gencheva-etal-2017-context} analogous to prioritization adopted by fact-checking organizations. With advent of transformers, fine-tuning based approaches that employ pre-trained transformer based language models as backbone was adopted for better performance \cite{stammbach-etal-2023-environmental,sheikhi-etal-2023-automated}.   More recently, Large Language Models have been employed for claim worthiness detection \cite{hyben2024multilingualmultitopicalbenchmarkfinetuned,vykopal-etal-2025-large,majer-snajder-2024-claim,dmonte2025claimverificationagelarge} in few-shot and zero-shot settings.  However, they underperform when compared to fine-tuned transformer-based classification approaches owing to subjectivity in checkworthiness detection and limitations of internal understanding of what constitutes check-worthy claims \cite{majer-snajder-2024-claim,amatya2026multilingualfactcheckingscalefinetuned,10.1145/3626772.3661361}. However, LLM-based predictions / annotations are not reliable due to poor confidence calibration \cite{si-etal-2024-large} and are also poorly calibrated with respect to human-based annotations. While verbalized confidence approaches, which prompt LLM to verbalize numerical confidence scores in text have been proposed, they exhibit confidence saturation \cite{wang2026calibrating}, where the LLM’s reported scores become
uninformative. They also suffer from overconfidence \cite{xu-etal-2025-language}.  To improve the reliability of LLM, AFaCTA \cite{ni-etal-2024-afacta} leverages self-consistency to calibrate the confidence of LLM predictions for claim worthiness. However, the calibration done based on self-consistency over multiple LLM-generated outputs could collapse to the wrong answer as they have high estimation error \cite{zhou2026a}. It also does not provide an indication of the calibrated confidence of LLMs in their predictions. Additionally, the approach does not ensure that the LLM predictions are calibrated to align with claim-worthiness notions adopted in human annotations. Conformal prediction~\cite{angelopoulos2023conformal} offers a complementary perspective by constructing set-valued prediction regions with marginal coverage guarantees, but does not relocate the point estimate; \name{} instead corrects the score itself using labelled neighbours.

\section{Method}
\label{sec:method}
\begin{algorithm}[H]
    \caption{\name{} based calibration for Check-Worthy Claim Detection}
    \label{alg:ares-claim}
    \begin{algorithmic}[1]
        \Require LLM claim check-worthiness predictor $J$, labeled set $\mathcal{L}$, unlabeled set $\mathcal{U}$, confidence level $1-\alpha$, calibration set size $k$
        \State Compute $\hat{c}_i = J(x_i)$ for all $i \in \mathcal{U} \cup \mathcal{L}$
        \ForAll{$x_i : x_i \in \mathcal{U}$}
            \State Retrieve $\mathcal{S}_i \subset \mathcal{L}$: the $k$ nearest neighbors of $x_i$ by semantic similarity
            \State Compute residuals $(Y_j - \hat{c}_j)$ for each $j \in \mathcal{S}_i$
            \State Compute calibrated score $\theta_i$ using Eq.~\ref{eqn:ppi-decomp}
        \EndFor
    \end{algorithmic}
    \label{alg:ppi}
\end{algorithm}
\name{} is a post-hoc calibration layer that corrects LLM check-worthiness predictions at inference time using a small labeled calibration set, without modifying the underlying model (Figure~\ref{fig:ppi_block_diagram_claim_checkworthiness}).

\begin{figure*}[t]
    \centering
    \resizebox{0.9\textwidth}{!}{%
    \begin{tikzpicture}[
        font=\small,
        box/.style={draw, rounded corners=3pt, align=center, minimum height=10mm, inner sep=6pt, semithick},
        model/.style={box, fill=blue!10, draw=blue!40!black},
        calib/.style={box, fill=orange!12, draw=orange!55!black},
        io/.style={box, fill=teal!10, draw=teal!45!black},
        outbox/.style={box, fill=green!12, draw=green!45!black},
        data/.style={cylinder, shape border rotate=90, aspect=0.12, draw=gray!55!black, fill=gray!8, align=center, inner sep=5pt, semithick},
        plusnode/.style={circle, draw=black, fill=yellow!30, inner sep=0.5pt, minimum size=5.5mm, semithick},
        arr/.style={-{Stealth[length=2.6mm]}, semithick},
        lab/.style={font=\scriptsize, fill=white, inner sep=1.5pt},
    ]
        \node[io] (claim) at (0,0) {Incoming claim $x_i$\\[1pt] \scriptsize ``80\% of GDP is spent\\[-2pt] \scriptsize on healthcare''};

        \node[model] (llm) at (4.7,1.5) {Frozen LLM / SLM\\ predictor $J$\\[1pt] \scriptsize (black box, no re-training)};

        \node[calib] (knn) at (4.7,-1.5) {$k$-NN retrieval\\[1pt] \scriptsize semantic similarity};
        \node[data] (store) at (1.2,-3.2) {Calibration set $\mathcal{L}$\\ \scriptsize human-labeled claims\\[-2pt] \scriptsize $(x_j,\, Y_j,\, \hat{c}_j)$};
        \node[calib] (res) at (9.2,-1.5) {Residual correction\\ $\bar{r}_i = \frac{1}{k}\sum_{j \in \mathcal{S}_i} \left( Y_j - \hat{c}_j \right)$};

        \node[plusnode] (plus) at (12.1,0) {$+$};
        \node[outbox] (out) at (14.5,0) {Calibrated score\\ $\theta_i = \hat{c}_i + \bar{r}_i \;\gtrless\; \epsilon$\\[1pt] \scriptsize check-worthy? + CI (Eq.~\ref{eqn:ppi-ci})};

        \begin{scope}[on background layer]
            \node[draw=orange!60!black, dashed, rounded corners=4pt, fill=orange!4,
                  fit=(knn) (store) (res), inner sep=7pt,
                  label={[font=\scriptsize\itshape, text=orange!50!black, anchor=south east]south east:\name{} calibration layer (post-hoc, at inference time)}] (caliblayer) {};
        \end{scope}

        \draw[arr] (claim.east) -- ++(0.5,0) |- (llm.west);
        \draw[arr] (claim.east) -- ++(0.5,0) |- (knn.west);
        \draw[arr] (store.east) -| (knn.south);
        \draw[arr] (knn.east) -- (res.west) node[lab, midway, above=3.5pt] {$\mathcal{S}_i$};
        \draw[arr] (llm.east) -| (plus.north) node[lab, pos=0.25, above=1pt] {$\hat{c}_i \in [0,1]$ \scriptsize(uncalibrated)};
        \draw[arr] (res.east) -| (plus.south) node[lab, pos=0.25, below=1pt] {$\bar{r}_i$};
        \draw[arr] (plus.east) -- (out.west);
    \end{tikzpicture}%
    }
    \caption{\footnotesize  \name{} overview. A frozen LM scores claim $x_i$ as $\hat{c}_i$; the $k$ nearest labeled neighbors
  $\mathcal{S}_i \subset \mathcal{L}$ provide a residual correction $\bar{r}_i$; the calibrated score
 $\theta_i = \hat{c}_i + \bar{r}_i$ is thresholded at $\epsilon$ to yield the final decision with a
per-instance CI.}
    \label{fig:ppi_block_diagram_claim_checkworthiness}
\end{figure*}
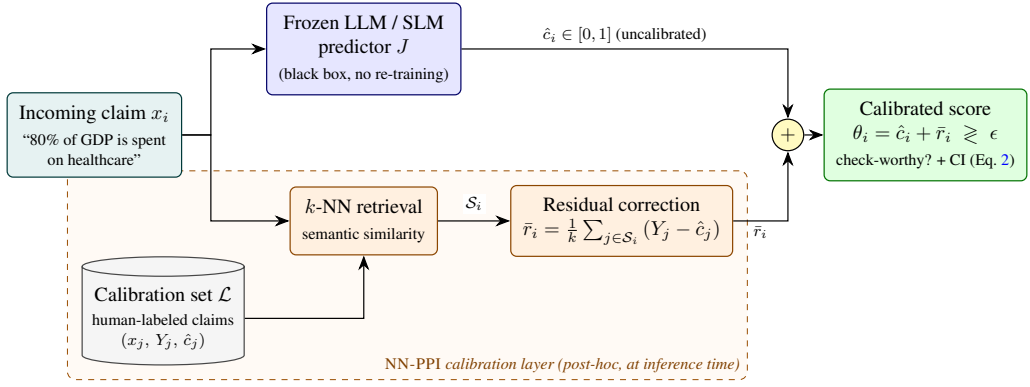

\begin{figure*}[t]
    \centering
    \begin{mdframed}[backgroundcolor=gray!20, linecolor=black, linewidth=0.5pt, innerleftmargin=8pt, innerrightmargin=8pt, innertopmargin=6pt, innerbottommargin=6pt]
        \footnotesize
        \setlength{\parskip}{0pt}
        \begin{ttfamily}
            \noindent \# YOUR ROLE 
            
            \noindent You are an impartial fact-checker. You are aware of what kind of statements that goes around news and published media are fact-check-worthy claims or not based on following check-worthiness criteria. 
            \noindent \begin{itemize}[leftmargin=1.8em, nosep, topsep=0pt, itemsep=1pt]
                \item High-stakes, society-level, quantitative or study-based claims are very highly check-worthy.
                \item Broad policy mechanism or sector-wide quantitative claims are highly check-worthy.
                \item Mid-tier, localized or mixed claims with numbers/opinions are medium check-worthy.
                \item Isolated incidents, hearsay, or loosely phrased generalizations are low check-worthy.
                \item Personal stories, greetings, meta, nostalgia, logistics are very low check-worthy.
                \item Statements containing factual claims that are not check-worthy by above check-worthiness criteria or non claim statements (e.g. opinions, speculations, feelings, rhetorical statements, campaign slogans, predictions) are not check-worthy.
            \end{itemize}
            \noindent \#\# Some examples for claim check-worthiness are below. 
            \noindent \{\{examples\}\}\\
            \noindent \# YOUR TASK
            
            \noindent You will be provided a statement which can be a claim or not. Make your best judgement using the claim check-worthiness criteria above and assign it a score of a value between 0 and 1 on how confident you are of it being a check-worthy claim. Assign it with higher confidence score if you are more confident about the statement being a claim, else assign it lower confidence score.
            \noindent Alongside claim check-worthiness confidence score, also try to provide a justification for your confidence score. Provide the justification in natural language and in no more than 100 words.
            \noindent \#\# Input statement: 
            
            \noindent \{\{claim\}\}
            \noindent \#\# Output format:
            \noindent Only output a JSON object with confidence\_score and justification that can be parsed by a JSON parser. Do not output any other text. Strictly format the output as JSON object below. 
            \noindent \{ \noindent \hspace*{0.5em} "confidence\_score": \textless{}a float value between 0 and 1\textgreater{}, \noindent \hspace*{0.5em} "justification": \textless{}a short natural language justification for the confidence\_score\textgreater{} \noindent \}
        \end{ttfamily}
    \end{mdframed}
    \caption{Prompt used for check-worthiness scoring with verbalized confidence from 0 to 1. The \texttt{\{\{examples\}\}} placeholder is filled with 6 fixed few-shot examples, one per check-worthiness tier (very high to not check-worthy).}
    \label{fig:check_worthiness_score_prompt}
\end{figure*}

\subsection{Problem Setup}

Prediction-Powered Inference (PPI)~\cite{angelopoulos2023predictionpoweredinference,angelopoulos2024ppiefficientpredictionpoweredinference} produces confidence intervals around \emph{population-level} metrics e.g., system-level accuracy of a RAG pipeline~\cite{saad-falcon-etal-2024-ares}, but provides no per-instance calibration. We extend PPI to the pointwise setting for claim check-worthiness, as illustrated in Figure~\ref{fig:ppi_block_diagram_claim_checkworthiness}.

Formally, let $\mathcal{U}$ be a large set of unlabeled claims and $\mathcal{L}$ a small human-labeled calibration set with binary labels $Y_j \in \{0,1\}$. An LLM predictor $J$ produces a continuous check-worthiness score $\hat{c}_i = J(x_i) \in [0,1]$ for each claim $x_i$, which is thresholded at $\epsilon{=}0.5$ to yield a binary prediction. Our goal is to compute a calibrated score $\theta_i$ for each test instance by leveraging residuals from the most semantically similar labeled instances (cosine similarity).

\subsection{Nearest Neighbor Prediction-Powered Inference (\name{}) Formulation}

To calibrate the LLM outputs we devise the \name{} algorithm as shown in Algorithm \ref{alg:ppi}. We employ
semantic similarity as the measure to select subset of the
calibration set using K-Nearest-Neighbor (KNN) for each test instance whose claim-worthiness is to be determined. Hence we retrieve $\mathcal{S}_i$, where $\mathcal{S}_i \subset \mathcal{L}$ consists of
labeled samples that are similar to the test sample $x_i$ from $\mathcal{U}$. We also evaluate
PPI setup under varying set size $k$ of the calibration set (KNN neighbor size). The final calibrated score $\theta_i$ can be obtained as follows:
\begin{equation}
    \theta_i = \hat{c}_i + \frac{1}{k}\sum_{j \in \mathcal{S}_i} \left( Y_j - \hat{c}_j \right)
    \label{eqn:ppi-decomp}
\end{equation}
\vspace{-1em}

The first term in above equation is the LLM check-worthiness prediction score for the test sentence,
while the second term corrects for residual bias estimated from the calibration subset.

Additionally, we also obtain a confidence interval for the calibrated score with $(1 - \alpha)$ confidence using residuals from the calibration set $\mathcal{S}_i$ as
$    \operatorname{Var}(\theta_{i}) \approx \frac{\sigma_{\mathrm{res}}^2}{|\mathcal{S}_i|},
     \text{where } \sigma_{\mathrm{res}}^2 = \operatorname{Var}_{j \in \mathcal{S}_i}(Y_j - \hat{c}_j).$
Thus, a $(1-\alpha)$ confidence interval is given by:
\begin{equation}
    \theta_i \pm z_{1-\alpha/2} \frac{\sigma_{\mathrm{res}}}{\sqrt{|\mathcal{S}_i|}}.
    \label{eqn:ppi-ci}
\end{equation}
The confidence intervals help gauge the uncertainty in LLM predictions.

This calibration approach provides:
(1) a PPI-inspired, residual-based correction of model predictions from a small human-labeled subset, and
(2) per-instance scores suitable for downstream decisions.

\section{Experimental Setup}
\label{sec:experimental_setup}
\begin{table}[ht!]
    \centering
    \small
    \begin{tabular}{lrrrr}
        \hline
        Dataset & Calib.~($|\mathcal{L}|$) & Test & CW & NCW \\
        \hline
        ClaimBuster & 1,314 {\scriptsize(of 2,487)} & 2,740 & 725  & 2,015 \\
        CLEF 2024   & 2,406 {\scriptsize(of 22,501)} & 317  & 107  & 210  \\
        \hline
    \end{tabular}
    \caption{\footnotesize Dataset statistics. Calib.\ = class-balanced subset of the training split used as $\mathcal{L}$ (fixed seed); sizes chosen for stable performance across all $k$ (Appendix~\ref{sec:ablation-labeled-set-size}). CW/NCW = check-worthy/not check-worthy counts in the test set.}
    \label{tab:claim_checkworthiness_datasets}
    \vspace{-1em}
\end{table}
\subsection{Dataset}
\label{sec:datasets}
The statistics on used datasets on claim check-worthiness are shown in Table \ref{tab:claim_checkworthiness_datasets}. \textbf{ClaimBuster} \cite{hassan-2017-claimbuster} dataset is a collection of 23,533 human-annotated statements extracted from all U.S. general election presidential debates held between 1960 and 2016. We use 2012 election data for calibration and 2016 data for testing. \textbf{CLEF 2024 - CheckThat! Task 1} \cite{alam-2021-fighting-covid} dataset is a multi-domain collection designed for claim check-worthiness detection across the languages: Arabic, Dutch, English, and Spanish. We limit our analysis for claim check-worthiness in english language. 
We construct $\mathcal{L}$ by class-balanced sampling from the train split of each dataset (fixed random seed), yielding 1,314 examples for ClaimBuster and 2,406 for CLEF 2024; an ablation (\ref{sec:ablation-labeled-set-size}) confirms this size is adequate.

\begin{table*}[hbt!]
    \centering
    \scriptsize
    \begin{tabular}{llrrrrrrrrr}
        \toprule
        & & \multicolumn{3}{c}{Weighted F1} & \multicolumn{3}{c}{Class 0 F1} & \multicolumn{3}{c}{Class 1 F1} \\
        \cmidrule(lr){3-5} \cmidrule(lr){6-8} \cmidrule(lr){9-11}
        Model & $k$ & \mbaseline & \mknn & \mknnppi & \mbaseline & \mknn & \mknnppi & \mbaseline & \mknn & \mknnppi \\
        \midrule
        \multicolumn{11}{c}{\textit{ClaimBuster 2016}} \\
        \midrule
        \multirow[t]{3}{*}{\claudeopus} & 3 & \textbf{0.832} & 0.703 & 0.816 & \textbf{0.905} & 0.780 & 0.873 & 0.631 & 0.490 & \textbf{0.658} \\
         & 5 & 0.832 & 0.720 & \textbf{0.836} & \textbf{0.905} & 0.794 & 0.890 & 0.631 & 0.512 & \textbf{0.685} \\
         & 10 & 0.832 & 0.706 & \textbf{0.858} & 0.905 & 0.767 & \textbf{0.906} & 0.631 & 0.539 & \textbf{0.723} \\
        \addlinespace
                \multirow[t]{3}{*}{\gptmodel} & 3 & \textbf{0.843} & 0.703 & 0.817 & \textbf{0.901} & 0.780 & 0.875 & \textbf{0.685} & 0.491 & 0.659 \\
         & 5 & \textbf{0.843} & 0.720 & 0.827 & \textbf{0.901} & 0.794 & 0.884 & \textbf{0.685} & 0.513 & 0.670 \\
         & 10 & 0.843 & 0.706 & \textbf{0.846} & \textbf{0.901} & 0.767 & 0.898 & 0.685 & 0.539 & \textbf{0.702} \\
         \addlinespace
         \midrule
         \addlinespace
         \multirow[t]{3}{*}{\xlmrobertalft} & 3 & \textbf{0.789} & 0.704 & 0.767 & \textbf{0.892} & 0.780 & 0.841 & 0.505 & 0.494 & \textbf{0.560} \\

         & 5 & 0.789 & 0.723 & \textbf{0.790} & \textbf{0.892} & 0.793 & 0.862 & 0.505 & 0.528 & \textbf{0.592} \\

         & 10 & 0.789 & 0.712 & \textbf{0.820} & \textbf{0.892} & 0.772 & 0.888 & 0.505 & 0.548 & \textbf{0.634} \\

         \addlinespace
         \midrule
                 \multirow[t]{3}{*}{\gemmasmall} & 3 & 0.114 & 0.697 & \textbf{0.698} & 0.000 & 0.773 & \textbf{0.774} & 0.425 & 0.491 & \textbf{0.492} \\
         & 5 & 0.114 & \textbf{0.710} & 0.709 & 0.000 & 0.783 & \textbf{0.784} & 0.425 & \textbf{0.509} & 0.506 \\
         & 10 & 0.114 & 0.703 & \textbf{0.721} & 0.000 & 0.764 & \textbf{0.790} & 0.425 & \textbf{0.539} & 0.532 \\
        \addlinespace
        \multirow[t]{3}{*}{\gemmaoneb} & 3 & 0.638 & 0.698 & \textbf{0.721} & 0.681 & 0.775 & \textbf{0.795} & \textbf{0.516} & 0.484 & 0.514 \\
         & 5 & 0.638 & 0.712 & \textbf{0.733} & 0.681 & 0.786 & \textbf{0.804} & 0.516 & 0.506 & \textbf{0.534} \\
         & 10 & 0.638 & 0.699 & \textbf{0.734} & 0.681 & 0.756 & \textbf{0.799} & 0.516 & 0.540 & \textbf{0.550} \\
        \addlinespace

        \multirow[t]{3}{*}{\gemmafourb} & 3 & 0.568 & 0.703 & \textbf{0.729} & 0.587 & 0.780 & \textbf{0.803} & 0.517 & 0.489 & \textbf{0.525} \\
         & 5 & 0.568 & 0.719 & \textbf{0.745} & 0.587 & 0.794 & \textbf{0.816} & 0.517 & 0.511 & \textbf{0.549} \\
         & 10 & 0.568 & 0.707 & \textbf{0.760} & 0.587 & 0.767 & \textbf{0.827} & 0.517 & 0.540 & \textbf{0.574} \\

        \midrule
        \multicolumn{11}{c}{\textit{CLEF 2024}} \\
        \midrule
        \multirow[t]{3}{*}{\claudeopus} & 3 & 0.855 & 0.790 & \textbf{0.866} & \textbf{0.903} & 0.831 & 0.895 & 0.761 & 0.709 & \textbf{0.809} \\
         & 5 & 0.855 & 0.801 & \textbf{0.899} & 0.903 & 0.844 & \textbf{0.925} & 0.761 & 0.717 & \textbf{0.848} \\
         & 10 & 0.855 & 0.804 & \textbf{0.899} & 0.903 & 0.834 & \textbf{0.925} & 0.761 & 0.744 & \textbf{0.848} \\
        \addlinespace
                \multirow[t]{3}{*}{\gptmodel} & 3 & 0.835 & 0.786 & \textbf{0.852} & \textbf{0.889} & 0.829 & 0.888 & 0.731 & 0.704 & \textbf{0.783} \\
         & 5 & 0.835 & 0.801 & \textbf{0.879} & 0.889 & 0.843 & \textbf{0.911} & 0.731 & 0.719 & \textbf{0.817} \\
         & 10 & 0.835 & 0.801 & \textbf{0.863} & 0.889 & 0.831 & \textbf{0.900} & 0.731 & 0.741 & \textbf{0.792} \\
         \addlinespace
         \midrule
         \addlinespace
         \multirow[t]{3}{*}{\xlmrobertalft} & 3 & 0.754 & 0.755 & \textbf{0.787} & \textbf{0.860} & 0.804 & 0.844 & 0.547 & 0.661 & \textbf{0.676} \\
         & 5 & 0.754 & 0.780 & \textbf{0.811} & 0.860 & 0.826 & \textbf{0.864} & 0.547 & 0.690 & \textbf{0.709} \\
         & 10 & 0.754 & 0.782 & \textbf{0.837} & 0.860 & 0.813 & \textbf{0.888} & 0.547 & 0.724 & \textbf{0.737} \\
         \addlinespace
         \midrule
                 \multirow[t]{3}{*}{\gemmasmall} & 3 & 0.179 & \textbf{0.801} & 0.800 & 0.000 & 0.843 & \textbf{0.844} & 0.513 & \textbf{0.721} & 0.719 \\
         & 5 & 0.179 & 0.824 & \textbf{0.827} & 0.000 & 0.859 & \textbf{0.863} & 0.513 & 0.759 & \textbf{0.760} \\
         & 10 & 0.179 & 0.809 & \textbf{0.822} & 0.000 & 0.843 & \textbf{0.864} & 0.513 & \textbf{0.747} & 0.745 \\
        \addlinespace
        \multirow[t]{3}{*}{\gemmaoneb} & 3 & 0.720 & \textbf{0.780} & 0.771 & 0.766 & 0.823 & \textbf{0.824} & 0.629 & \textbf{0.695} & 0.667 \\
         & 5 & 0.720 & \textbf{0.801} & 0.777 & 0.766 & \textbf{0.842} & 0.829 & 0.629 & \textbf{0.719} & 0.676 \\
         & 10 & 0.720 & \textbf{0.803} & 0.761 & 0.766 & \textbf{0.834} & 0.817 & 0.629 & \textbf{0.742} & 0.651 \\
        \addlinespace

        \multirow[t]{3}{*}{\gemmafourb} & 3 & 0.688 & 0.790 & \textbf{0.795} & 0.706 & 0.831 & \textbf{0.838} & 0.655 & 0.709 & \textbf{0.711} \\
         & 5 & 0.688 & 0.801 & \textbf{0.824} & 0.706 & 0.843 & \textbf{0.866} & 0.655 & 0.719 & \textbf{0.743} \\
         & 10 & 0.688 & 0.807 & \textbf{0.827} & 0.706 & 0.837 & \textbf{0.869} & 0.655 & \textbf{0.747} & 0.744 \\
        \addlinespace

        \bottomrule
    \end{tabular}
    \caption{Weighted F1 for \mbaseline{} (few-shot), \mknn{}, and \mknnppi{} with \# of calibration samples $k=3,5,10$.}
    \label{tab:ppi_weighted_scores}
    \vspace{-2em}
\end{table*}

\subsection{Models}
We use few-shot prompting to produce a check-worthiness confidence score in $[0,1]$ (prompt in Figure~\ref{fig:check_worthiness_score_prompt}) across three model classes: small language models (SLMs, $\leq$4B parameters) served via Ollama, a fine-tuned XLM-RoBERTa-Large model in production at an early-stage startup (Appendix \ref{sec:xlm_roberta_ft}), and large commercial API models (GPT-5.2 and Claude Opus 4.6).\footnote{Both accessed via Azure AI endpoints: \texttt{gpt-5.2} (version 2025-03-01-preview, knowledge cutoff Aug 2025) and \texttt{claude-opus-4-6} (knowledge cutoff May 2025).} The parameters used for experiments are in Appendix \ref{sec:model-parameters}.

\subsection{\name{} Implementation Details}
We apply \name{} based calibration to adjust baseline predictions. A manually annotated calibration set is indexed in a ChromaDB vector store using \texttt{all-MiniLM-L6-v2} sentence embeddings~\cite{reimers2019sentencebertsentenceembeddingsusing} with cosine similarity. For each test claim, we retrieve the $k$ nearest neighbors to form a localized calibration set. The baseline LLM prediction is then adjusted using this set (Equation \ref{eqn:ppi-decomp}), and converted to a binary label using a threshold $\epsilon = 0.5$.

\noindent \textbf{Evaluation Metrics}:
We report weighted F1, classwise F1 (check-worthy and non-check-worthy). %

\section{Results}
\begin{table*}[t]
    \centering
    \footnotesize
    \begin{tabular}{lp{5.8cm}ccc}
        \toprule
        Failure mode & Claim & Gold & Base & \name{} \\
        \midrule
        \multicolumn{5}{l}{\textit{Fixed (corrected by \name{})}} \\
        \addlinespace[2pt]
        FP $\to$ correct
            & \textit{``It has been the policy of the United States, Democrats and Republicans, to do everything we can\ldots''}
            & 0 & 0.80 & 0.00 \\
        \addlinespace[2pt]
        FP $\to$ correct
            & \textit{``Mental health is one of the biggest concerns, because now police are having to handle a lot of situations\ldots''}
            & 0 & 0.80 & 0.03 \\
        \addlinespace[4pt]
        FN $\to$ correct
            & \textit{``President Obama said you keep your doctor, you keep your plan.''}
            & 1 & 0.10 & 0.80 \\
        \addlinespace[2pt]
        FN $\to$ correct
            & \textit{``Iran now and Russia are now against us.''}
            & 1 & 0.10 & 0.52 \\
        \midrule
        \multicolumn{5}{l}{\textit{Persistent (not corrected)}} \\
        \addlinespace[2pt]
        FP persist
            & \textit{``I was in the Senate before I became secretary of state.''}
            & 0 & 0.60 & 1.03 \\
        \addlinespace[2pt]
        FN persist
            & \textit{``I have no loans from Russia.''}
            & 1 & 0.30 & $-$0.33 \\
        \midrule
        \multicolumn{5}{l}{\textit{Regression (new error introduced)}} \\
        \addlinespace[2pt]
        Correct $\to$ FN
            & \textit{``We have to protect our inner cities, because African-American communities are being decimated\ldots''}
            & 1 & 0.80 & 0.07 \\
        \addlinespace[2pt]
        Correct $\to$ FN
            & \textit{``But the Middle East still controls a lot of the prices.''}
            & 1 & 0.70 & $-$0.03 \\
        \bottomrule
    \end{tabular}
    \caption{Qualitative failure modes of \gemmafourb{} on ClaimBuster ($k{=}3$). Base and \name{} columns show the raw calibrated score (threshold $\epsilon{=}0.5$). Scores outside $[0,1]$ arise because the residual correction is unconstrained.}
    \label{tab:error_analysis}
\end{table*}

\begin{table}[t]
    \centering
    \small
    \begin{tabular}{p{0.3cm}rrrrrr}
        \toprule
        & \multicolumn{3}{c}{ClaimBuster} & \multicolumn{3}{c}{CLEF 2024} \\
        \cmidrule(lr){2-4} \cmidrule(lr){5-7}
        $k$ & Cls-0 & Cls-1 & Overall & Cls-0 & Cls-1 & Overall \\
        \midrule
        3  & 68.8 & 53.6 & 64.8 & 72.7 & 60.5 & 68.5 \\
        5  & 66.3 & 46.8 & 61.1 & 69.5 & 56.8 & 65.2 \\
        10 & 51.6 & 32.8 & 46.6 & 58.1 & 42.1 & 52.6 \\
        \bottomrule
    \end{tabular}
    \caption{\footnotesize Empirical 95\% CI coverage (\%) averaged across models. Cls-1 (check-worthy) consistently under-covers more than Cls-0; smaller $k$ yields the best-calibrated intervals.}
    \vspace{-0.5cm}
    \label{tab:ci_coverage}
\end{table}

To answer \textbf{RQ 1} and \textbf{RQ 2}, we compare three conditions: (1) few-shot LLM prediction with no calibration (Baseline), (2) KNN label averaging without the PPI correction, and (3) \name{} calibration. Results are shown in Table~\ref{tab:ppi_weighted_scores} for $k \in \{3, 5, 10\}$.
\name{} improves weighted F1 across all model scales (\textbf{RQ 1}), with gains largest for smaller LLMs. We observe the most significant gains for \gemmasmall{}: its baseline achieves only 0.114 on ClaimBuster (near-random due to severe class prediction bias) and 0.179 on CLEF, yet when calibrated with \name{}, it achieves a competitive 0.721 and 0.827, respectively. Mid-scale SLMs also benefit substantially: \gemmafourb{} gains \textbf{+33.80\%} on ClaimBuster (0.568\,$\to$\,0.760) and \textbf{+20\%} on CLEF (0.688\,$\to$\,0.827), while \gemmaoneb{} gains \textbf{+15\%} and \textbf{+8\%}. Larger models are mostly saturated: \claudeopus{} and \gptmodel{} each gain at most 5\% on CLEF and near zero or slightly negative on ClaimBuster, where baselines already exceed 0.83. 

We also calibrate \xlmrobertalft{}, a fine-tuned model from the production pipeline
(Table~\ref{tab:ppi_weighted_scores}). \name{} yields gains of \textbf{11\%} over the baseline and \textbf{7.03\%} over KNN on CLEF 2024. Despite being fine-tuned on in-domain data, \xlmrobertalft{} ranks second only to frontier LLMs, confirming that residual calibration is complementary to supervised fine-tuning.

Addressing \textbf{RQ 2}, \name{} outperforms the plain KNN baseline in the majority of conditions. The exception is \gemmaoneb{} on CLEF, where KNN matches or slightly exceeds \name{} (0.803 vs.\ 0.761 at $k{=}10$). The PPI residual correction is most beneficial when the model's bias is systematic. Hence, for a reasonably well-calibrated model, KNN averaging alone may suffice.

One observation worth clarifying: \gemmafourb{} scores a lower baseline weighted F1 than \gemmaoneb{} on ClaimBuster (0.568 vs.\ 0.638) despite being larger. This is a class-imbalance artifact: \gemmafourb{} over-predicts the positive class (65.7\% vs.\ a true rate of 26.5\%), which collapses cls-0 F1 and dominates weighted F1 in this imbalanced dataset (see Table~\ref{tab:baseline_analysis}, Appendix~\ref{sec:baseline_analysis}). \name{} corrects this bias, which is why \gemmafourb{} benefits more from calibration (+34\%) than \gemmaoneb{} (+15\%).

To answer \textbf{RQ3}, we analyze the results in Table \ref{tab:ppi_weighted_scores} and we observe that both classwise and weighted $F1$ scores saturate as we advance from k=3 to 5, 10. The possible explanation for bigger neighbor size saturating calibration performance could be introduction of noise in residuals by non representative samples. As distributionally dis-similar examples get included in the calibration set, it increases the uncertainty of the calibration procedure. 

Hyperparameters for all models are reported in Appendix~\ref{sec:model-parameters}. We also evaluate \name{} at a lower temperature ($T{=}0.1$) in Appendix~\ref{sec:nn-ppi-temp-0.1}. The key findings hold across both settings; \gptmodel{} shows a slight baseline improvement at lower temperature due to more deterministic sampling, while \gemmaoneb{} and \gemmafourb{} are unaffected, as their prediction biases are structural rather than sampling-induced.

\paragraph{Failure mode analysis.}
Table~\ref{tab:error_analysis} shows four failure modes for \gemmafourb{} on ClaimBuster. \name{} succeeds by pulling over-triggered rhetorical claims below threshold (\textbf{fixed FPs}) and rescuing missed factual claims via high-label neighbors (\textbf{fixed FNs}). It fails when the neighborhood is itself biased (\textbf{persistent errors}) or when topically unrelated neighbors overcorrect a borderline prediction (\textbf{regressions}).

\textbf{Confidence interval analysis.}
Table~\ref{tab:ci_coverage} reports empirical CI coverage for ClaimBuster and CLEF 2024 datasets. Firstly, we observe that empirical coverage falls as we increase the number of neighbors considered (moving from $k=3$ to $k=10$. We also observe that  CIs-1 for class 1 in ClaimBuster has lower coverage even at $k=3$, which explains the low performance of different models across approaches as observed from classwise F1. However, for CLEF 2024, we observe that \name{} achieves a balanced performance on both classes compared to performance on ClaimBuster. We observe that this is primarily because coverage for both classes is better in CLEF 2024 compared to ClaimBuster.
Smaller $k$ yields better-calibrated intervals (64.8 overall coverage on Claimbuster at $k=3$ vs 46.6 with $k=10$ at 95\%), since tighter neighborhoods produce more homogeneous residuals. Second, decision accuracy (correct side of the $\epsilon{=}0.5$ threshold) remains high at 70--90\% across models and datasets.  CIs can be treated as relative uncertainty indicators than frequentist guarantees.

\section{Conclusion}
\label{sec:conclusion}

We propose \name{}, a pointwise approach for calibrating LLM predictions in claim check-worthiness detection. Our approach leverages a small, human-annotated calibration set and nearest-neighbor residual correction.
This results in the largest performance gains for SLMs, narrowing the gap with larger models. We also observe similar gains for our smaller transformer models deployed in production, demonstrating the efficiency of the proposed approach.  In the future, we also plan to tackle the \emph{adaptivity dimension}, where the calibration set can be updated to adapt to new notions of claim worthiness which may evolve over time without any underlying model changes.

\newpage
\section{Limitations}
\label{sec:limitations}

\paragraph{Relation to standard calibration.}
Parametric post-hoc calibrators such as Platt scaling, temperature scaling, and isotonic regression fit a single global score-to-probability mapping, and are most effective when miscalibration is homogeneous across the input space. \name{} instead applies a local, non-parametric correction driven by labelled neighbours, targeting input-dependent bias (e.g., topic-specific over-prediction) that a global monotone map cannot capture. Conformal prediction, relatedly, yields marginally valid prediction sets but does not relocate the point estimate. A controlled comparison against these methods is the natural next step; our focus here is the deployment question of recovering LLM-level decisions from cheap models.

Our proposed approach primarily tackles calibrating LLM predictions to align closer to human judgments, using a statistically sound PPI-inspired approach, and works well for black-box API models and open-source models. Our proposed approach relies on a small yet distributionally representative calibration set.  While this requires human annotations or repurposing of existing training sets, it is minimal effort and provides the advantage of principled calibration. However, optimally selecting a subset of calibration samples that are distributionally similar to the test sample is a challenge. While semantic similarity works well as a proxy in our experiments, semantic relevance may not always translate to distributional similarity. While one could explore Wasserstein-based distributional similarity measures, they are quite computationally intensive and hence beyond scope as our focus is on lightweight post-hoc calibration of language model outputs. In the future, we plan to explore alternative efficient data selection mechanisms for dynamically selecting from the calibration set.

\section{Ethical Considerations and Risks}

The datasets we use in this work are drawn from public sources (with Creative Commons license) and include no personally identifiable or sensitive information. The claims focus on public data and domain-specific knowledge rather than private individuals. Our approach primarily focuses on calibrating LLM predictions to make them more trustworthy, as they are prone to hallucinations and uncalibrated confidence in their predictions. Our approach does not currently include fairness or bias mitigation across demographic attributes, which may be relevant for politically or socially sensitive claims, but is beyond the scope of the focused contribution on calibration in this work.

\bibliography{00paper}

@article{world_economic_forum,
  address =       {New York, NY, USA},
  author =        {Webb, Helena and Jirotka, Marina and
                   Stahl, Bernd Carsten and Housley, William and
                   Edwards, Adam and Williams, Matthew and Procter, Rob and
                   Rana, Omer and Burnap, Pete},
  journal =       {SIGCAS Comput. Soc.},
  number =        {3},
  pages =         {193–201},
  publisher =     {Association for Computing Machinery},
  title =         {Digital wildfires: hyper-connectivity, havoc and a
                   global ethos to govern social media},
  volume =        {45},
  year =          {2016},
  doi =           {10.1145/2874239.2874267},
  issn =          {0095-2737},
  url =           {https://doi.org/10.1145/2874239.2874267},
}

@inproceedings{stammbach-etal-2023-environmental,
  address =       {Toronto, Canada},
  author =        {Stammbach, Dominik and Webersinke, Nicolas and
                   Bingler, Julia and Kraus, Mathias and
                   Leippold, Markus},
  booktitle =     {Proceedings of the 61st Annual Meeting of the
                   Association for Computational Linguistics (Volume 2:
                   Short Papers)},
  editor =        {Rogers, Anna and Boyd-Graber, Jordan and
                   Okazaki, Naoaki},
  pages =         {1051--1066},
  publisher =     {Association for Computational Linguistics},
  title =         {Environmental Claim Detection},
  year =          {2023},
  doi =           {10.18653/v1/2023.acl-short.91},
  url =           {https://aclanthology.org/2023.acl-short.91/},
}

@inproceedings{sheikhi-etal-2023-automated,
  address =       {T{\'o}rshavn, Faroe Islands},
  author =        {Sheikhi, Ghazaal and Touileb, Samia and Khan, Sohail},
  booktitle =     {Proceedings of the 24th Nordic Conference on
                   Computational Linguistics (NoDaLiDa)},
  editor =        {Alum{\"a}e, Tanel and Fishel, Mark},
  pages =         {1--9},
  publisher =     {University of Tartu Library},
  title =         {Automated Claim Detection for Fact-checking: A Case
                   Study using {N}orwegian Pre-trained Language Models},
  year =          {2023},
  url =           {https://aclanthology.org/2023.nodalida-1.1/},
}

@misc{hyben2024multilingualmultitopicalbenchmarkfinetuned,
  author =        {Martin Hyben and Sebastian Kula and Ivan Srba and
                   Robert Moro and Jakub Simko},
  title =         {Multilingual and Multi-topical Benchmark of
                   Fine-tuned Language models and Large Language Models
                   for Check-Worthy Claim Detection},
  year =          {2023},
  url =           {https://arxiv.org/abs/2311.06121},
}

@inproceedings{majer-snajder-2024-claim,
  address =       {Miami, Florida, USA},
  author =        {Majer, Laura and {\v{S}}najder, Jan},
  booktitle =     {Proceedings of the Seventh Fact Extraction and
                   VERification Workshop (FEVER)},
  editor =        {Schlichtkrull, Michael and Chen, Yulong and
                   Whitehouse, Chenxi and Deng, Zhenyun and
                   Akhtar, Mubashara and Aly, Rami and Guo, Zhijiang and
                   Christodoulopoulos, Christos and Cocarascu, Oana and
                   Mittal, Arpit and Thorne, James and Vlachos, Andreas},
  pages =         {245--263},
  publisher =     {Association for Computational Linguistics},
  title =         {Claim Check-Worthiness Detection: How Well do {LLM}s
                   Grasp Annotation Guidelines?},
  year =          {2024},
  doi =           {10.18653/v1/2024.fever-1.27},
  url =           {https://aclanthology.org/2024.fever-1.27/},
}

@inproceedings{ni-etal-2024-afacta,
  address =       {Bangkok, Thailand},
  author =        {Ni, Jingwei and Shi, Minjing and Stammbach, Dominik and
                   Sachan, Mrinmaya and Ash, Elliott and
                   Leippold, Markus},
  booktitle =     {Proceedings of the 62nd Annual Meeting of the
                   Association for Computational Linguistics (Volume 1:
                   Long Papers)},
  editor =        {Ku, Lun-Wei and Martins, Andre and Srikumar, Vivek},
  pages =         {1890--1912},
  publisher =     {Association for Computational Linguistics},
  title =         {{AF}a{CTA}: Assisting the Annotation of Factual Claim
                   Detection with Reliable {LLM} Annotators},
  year =          {2024},
  doi =           {10.18653/v1/2024.acl-long.104},
  url =           {https://aclanthology.org/2024.acl-long.104/},
}

@inproceedings{si-etal-2024-large,
  address =       {Mexico City, Mexico},
  author =        {Si, Chenglei and Goyal, Navita and Wu, Tongshuang and
                   Zhao, Chen and Feng, Shi and Daum{\'e} III, Hal and
                   Boyd-Graber, Jordan},
  booktitle =     {Proceedings of the 2024 Conference of the North
                   American Chapter of the Association for Computational
                   Linguistics: Human Language Technologies (Volume 1:
                   Long Papers)},
  editor =        {Duh, Kevin and Gomez, Helena and Bethard, Steven},
  pages =         {1459--1474},
  publisher =     {Association for Computational Linguistics},
  title =         {Large Language Models Help Humans Verify Truthfulness
                   {--} Except When They Are Convincingly Wrong},
  year =          {2024},
  doi =           {10.18653/v1/2024.naacl-long.81},
  url =           {https://aclanthology.org/2024.naacl-long.81/},
}

@inproceedings{zhuo-etal-2024-prosa,
  address =       {Miami, Florida, USA},
  author =        {Zhuo, Jingming and Zhang, Songyang and Fang, Xinyu and
                   Duan, Haodong and Lin, Dahua and Chen, Kai},
  booktitle =     {Findings of the Association for Computational
                   Linguistics: EMNLP 2024},
  editor =        {Al-Onaizan, Yaser and Bansal, Mohit and
                   Chen, Yun-Nung},
  pages =         {1950--1976},
  publisher =     {Association for Computational Linguistics},
  title =         {{P}ro{SA}: Assessing and Understanding the Prompt
                   Sensitivity of {LLM}s},
  year =          {2024},
  doi =           {10.18653/v1/2024.findings-emnlp.108},
  url =           {https://aclanthology.org/2024.findings-emnlp.108/},
}

@misc{angelopoulos2023predictionpoweredinference,
  author =        {Anastasios N. Angelopoulos and Stephen Bates and
                   Clara Fannjiang and Michael I. Jordan and
                   Tijana Zrnic},
  title =         {Prediction-Powered Inference},
  year =          {2023},
  url =           {https://arxiv.org/abs/2301.09633},
}

@misc{angelopoulos2024ppiefficientpredictionpoweredinference,
  author =        {Anastasios N. Angelopoulos and John C. Duchi and
                   Tijana Zrnic},
  title =         {PPI++: Efficient Prediction-Powered Inference},
  year =          {2023},
  url =           {https://arxiv.org/abs/2311.01453},
}

@inproceedings{supervised_CW,
  author =        {Naeemul Hassan and Fatma Arslan and Chengkai Li and
                   Mark Tremayne},
  booktitle =     {Proceedings of the 23rd {ACM} {SIGKDD} International
                   Conference on Knowledge Discovery and Data Mining,
                   Halifax, NS, Canada, August 13 - 17, 2017},
  pages =         {1803--1812},
  publisher =     {{ACM}},
  title =         {Toward Automated Fact-Checking: Detecting
                   Check-worthy Factual Claims by ClaimBuster},
  year =          {2017},
  bibsource =     {dblp computer science bibliography, https://dblp.org},
  doi =           {10.1145/3097983.3098131},
  url =           {https://doi.org/10.1145/3097983.3098131},
}

@inproceedings{wright-augenstein-2020-claim,
  address =       {Online},
  author =        {Wright, Dustin and Augenstein, Isabelle},
  booktitle =     {Findings of the Association for Computational
                   Linguistics: EMNLP 2020},
  editor =        {Cohn, Trevor and He, Yulan and Liu, Yang},
  pages =         {476--488},
  publisher =     {Association for Computational Linguistics},
  title =         {Claim Check-Worthiness Detection as Positive
                   Unlabelled Learning},
  year =          {2020},
  doi =           {10.18653/v1/2020.findings-emnlp.43},
  url =           {https://aclanthology.org/2020.findings-emnlp.43/},
}

@inproceedings{gencheva-etal-2017-context,
  address =       {Varna, Bulgaria},
  author =        {Gencheva, Pepa and Nakov, Preslav and
                   M{\`a}rquez, Llu{\'i}s and
                   Barr{\'o}n-Cede{\~n}o, Alberto and Koychev, Ivan},
  booktitle =     {Proceedings of the International Conference Recent
                   Advances in Natural Language Processing, {RANLP}
                   2017},
  editor =        {Mitkov, Ruslan and Angelova, Galia},
  pages =         {267--276},
  publisher =     {INCOMA Ltd.},
  title =         {A Context-Aware Approach for Detecting Worth-Checking
                   Claims in Political Debates},
  year =          {2017},
  doi =           {10.26615/978-954-452-049-6_037},
  url =           {https://aclanthology.org/R17-1037/},
}

@inproceedings{jaradat-etal-2018-claimrank,
  address =       {New Orleans, Louisiana},
  author =        {Jaradat, Israa and Gencheva, Pepa and
                   Barr{\'o}n-Cede{\~n}o, Alberto and
                   M{\`a}rquez, Llu{\'i}s and Nakov, Preslav},
  booktitle =     {Proceedings of the 2018 Conference of the North
                   {A}merican Chapter of the Association for
                   Computational Linguistics: Demonstrations},
  editor =        {Liu, Yang and Paek, Tim and Patwardhan, Manasi},
  pages =         {26--30},
  publisher =     {Association for Computational Linguistics},
  title =         {{C}laim{R}ank: Detecting Check-Worthy Claims in
                   {A}rabic and {E}nglish},
  year =          {2018},
  doi =           {10.18653/v1/N18-5006},
  url =           {https://aclanthology.org/N18-5006/},
}

@inproceedings{vykopal-etal-2025-large,
  address =       {Suzhou, China},
  author =        {Vykopal, Ivan and Pikuliak, Mat{\'u}{\v{s}} and
                   Ostermann, Simon and Anikina, Tatiana and
                   Gregor, Michal and Simko, Marian},
  booktitle =     {Findings of the Association for Computational
                   Linguistics: EMNLP 2025},
  editor =        {Christodoulopoulos, Christos and Chakraborty, Tanmoy and
                   Rose, Carolyn and Peng, Violet},
  pages =         {15741--15765},
  publisher =     {Association for Computational Linguistics},
  title =         {Large Language Models for Multilingual Previously
                   Fact-Checked Claim Detection},
  year =          {2025},
  doi =           {10.18653/v1/2025.findings-emnlp.852},
  isbn =          {979-8-89176-335-7},
  url =           {https://aclanthology.org/2025.findings-emnlp.852/},
}

@inproceedings{dmonte2025claimverificationagelarge,
  address =       {San Diego, California, United States},
  author =        {Dmonte, Alphaeus and Oruche, Roland R and
                   Zampieri, Marcos and Calyam, Prasad and
                   Augenstein, Isabelle},
  booktitle =     {Proceedings of the 64th Annual Meeting of the
                   {A}ssociation for {C}omputational {L}inguistics
                   (Volume 4: Student Research Workshop)},
  editor =        {T.Y.S.S., Santosh and Rodriguez, Juan Diego and
                   de Gibert, Ona},
  pages =         {15--29},
  publisher =     {Association for Computational Linguistics},
  title =         {Claim Verification in the Age of Large Language
                   Models: A Survey},
  year =          {2026},
  doi =           {10.18653/v1/2026.acl-srw.2},
  isbn =          {979-8-89176-393-7},
  url =           {https://aclanthology.org/2026.acl-srw.2/},
}

@misc{amatya2026multilingualfactcheckingscalefinetuned,
  author =        {Pratuat Amatya and Vinay Setty},
  title =         {Multilingual Fact-Checking at Scale: Fine-Tuned
                   Compact Models vs LLMs},
  year =          {2026},
  url =           {https://arxiv.org/abs/2606.08605},
}

@inproceedings{10.1145/3626772.3661361,
  author =        {Vinay Setty},
  booktitle =     {Proceedings of the 47th International {ACM} {SIGIR}
                   Conference on Research and Development in Information
                   Retrieval, {SIGIR} 2024, Washington DC, USA, July
                   14-18, 2024},
  editor =        {Grace Hui Yang and Hongning Wang and Sam Han and
                   Claudia Hauff and Guido Zuccon and Yi Zhang},
  pages =         {2842--2846},
  publisher =     {{ACM}},
  title =         {Surprising Efficacy of Fine-Tuned Transformers for
                   Fact-Checking over Larger Language Models},
  year =          {2024},
  bibsource =     {dblp computer science bibliography, https://dblp.org},
  doi =           {10.1145/3626772.3661361},
  url =           {https://doi.org/10.1145/3626772.3661361},
}

@inproceedings{wang2026calibrating,
  author =        {Victor Wang and Elias Stengel-Eskin},
  booktitle =     {The Fourteenth International Conference on Learning
                   Representations},
  title =         {Calibrating Verbalized Confidence with Self-Generated
                   Distractors},
  year =          {2026},
  url =           {https://openreview.net/forum?id=pZs4hhemXc},
}

@inproceedings{xu-etal-2025-language,
  address =       {Vienna, Austria},
  author =        {Xu, Chenjun and Wen, Bingbing and Han, Bin and
                   Wolfe, Robert and Wang, Lucy Lu and Howe, Bill},
  booktitle =     {Findings of the Association for Computational
                   Linguistics: ACL 2025},
  editor =        {Che, Wanxiang and Nabende, Joyce and
                   Shutova, Ekaterina and Pilehvar, Mohammad Taher},
  pages =         {25655--25672},
  publisher =     {Association for Computational Linguistics},
  title =         {Do Language Models Mirror Human Confidence? Exploring
                   Psychological Insights to Address Overconfidence in
                   {LLM}s},
  year =          {2025},
  doi =           {10.18653/v1/2025.findings-acl.1316},
  isbn =          {979-8-89176-256-5},
  url =           {https://aclanthology.org/2025.findings-acl.1316/},
}

@inproceedings{zhou2026a,
  author =        {Zhi Zhou and Tan Yuhao and Zenan Li and Yuan Yao and
                   Lan{-}Zhe Guo and Yufeng Li and Xiaoxing Ma},
  booktitle =     {Advances in Neural Information Processing Systems 38:
                   Annual Conference on Neural Information Processing
                   Systems 2025, NeurIPS 2025, San Diego, CA, USA,
                   December 2-7, 2025 / Mexico City, Mexico, November 30
                   - December 5, 2025},
  editor =        {Danielle Belgrave and Cheng Zhang and
                   Laura N. Montoya and Hsuan{-}Tien Lin and
                   Razvan Pascanu and Piotr Koniusz and Marzyeh Ghassemi and
                   Nancy Chen and Iv{\'{a}}n Vladimir Meza Ru{\'{\i}}z and
                   Arturo Loaiza{-}Bonilla},
  title =         {A Theoretical Study on Bridging Internal Probability
                   and Self-Consistency for {LLM} Reasoning},
  year =          {2025},
  bibsource =     {dblp computer science bibliography, https://dblp.org},
  url =           {http://papers.nips.cc/paper\_files/paper/2025/hash/
                  7e9afa9a02857bce4515247842471444-Abstract-Conference.html},
}

@article{angelopoulos2023conformal,
  author =        {Angelopoulos, Anastasios N and Bates, Stephen},
  journal =       {Foundations and Trends in Machine Learning},
  number =        {4},
  pages =         {494--591},
  publisher =     {Emerald Publishing Limited},
  title =         {Conformal prediction: A gentle introduction},
  volume =        {16},
  year =          {2023},
}

@inproceedings{saad-falcon-etal-2024-ares,
  address =       {Mexico City, Mexico},
  author =        {Saad-Falcon, Jon and Khattab, Omar and
                   Potts, Christopher and Zaharia, Matei},
  booktitle =     {Proceedings of the 2024 Conference of the North
                   American Chapter of the Association for Computational
                   Linguistics: Human Language Technologies (Volume 1:
                   Long Papers)},
  editor =        {Duh, Kevin and Gomez, Helena and Bethard, Steven},
  pages =         {338--354},
  publisher =     {Association for Computational Linguistics},
  title =         {{ARES}: An Automated Evaluation Framework for
                   Retrieval-Augmented Generation Systems},
  year =          {2024},
  doi =           {10.18653/v1/2024.naacl-long.20},
  url =           {https://aclanthology.org/2024.naacl-long.20/},
}

@article{hassan-2017-claimbuster,
  author =        {Hassan, Naeemul and Zhang, Gensheng and Arslan, Fatma and
                   Caraballo, Josue and Jimenez, Damian and
                   Gawsane, Siddhant and Hasan, Shohedul and
                   Joseph, Minumol and Kulkarni, Aaditya and
                   Nayak, Anil Kumar and Sable, Vikas and Li, Chengkai and
                   Tremayne, Mark},
  journal =       {Proc. VLDB Endow.},
  number =        {12},
  pages =         {1945–1948},
  publisher =     {VLDB Endowment},
  title =         {ClaimBuster: the first-ever end-to-end fact-checking
                   system},
  volume =        {10},
  year =          {2017},
  doi =           {10.14778/3137765.3137815},
  issn =          {2150-8097},
  url =           {https://doi.org/10.14778/3137765.3137815},
}

@inproceedings{alam-2021-fighting-covid,
  address =       {Punta Cana, Dominican Republic},
  author =        {Alam, Firoj and Shaar, Shaden and Dalvi, Fahim and
                   Sajjad, Hassan and Nikolov, Alex and Mubarak, Hamdy and
                   Da San Martino, Giovanni and Abdelali, Ahmed and
                   Durrani, Nadir and Darwish, Kareem and
                   Al-Homaid, Abdulaziz and Zaghouani, Wajdi and
                   Caselli, Tommaso and Danoe, Gijs and Stolk, Friso and
                   Bruntink, Britt and Nakov, Preslav},
  booktitle =     {Findings of the Association for Computational
                   Linguistics: EMNLP 2021},
  editor =        {Moens, Marie-Francine and Huang, Xuanjing and
                   Specia, Lucia and Yih, Scott Wen-tau},
  pages =         {611--649},
  publisher =     {Association for Computational Linguistics},
  title =         {Fighting the {COVID}-19 Infodemic: Modeling the
                   Perspective of Journalists, Fact-Checkers, Social
                   Media Platforms, Policy Makers, and the Society},
  year =          {2021},
  doi =           {10.18653/v1/2021.findings-emnlp.56},
  url =           {https://aclanthology.org/2021.findings-emnlp.56/},
}

@inproceedings{reimers2019sentencebertsentenceembeddingsusing,
  address =       {Hong Kong, China},
  author =        {Reimers, Nils and Gurevych, Iryna},
  booktitle =     {Proceedings of the 2019 Conference on Empirical
                   Methods in Natural Language Processing and the 9th
                   International Joint Conference on Natural Language
                   Processing (EMNLP-IJCNLP)},
  editor =        {Inui, Kentaro and Jiang, Jing and Ng, Vincent and
                   Wan, Xiaojun},
  pages =         {3982--3992},
  publisher =     {Association for Computational Linguistics},
  title =         {Sentence-{BERT}: Sentence Embeddings using {S}iamese
                   {BERT}-Networks},
  year =          {2019},
  doi =           {10.18653/v1/D19-1410},
  url =           {https://aclanthology.org/D19-1410/},
}

@inproceedings{conneau-etal-2020-unsupervised,
  address =       {Online},
  author =        {Conneau, Alexis and Khandelwal, Kartikay and
                   Goyal, Naman and Chaudhary, Vishrav and
                   Wenzek, Guillaume and Guzm{\'a}n, Francisco and
                   Grave, Edouard and Ott, Myle and Zettlemoyer, Luke and
                   Stoyanov, Veselin},
  booktitle =     {Proceedings of the 58th Annual Meeting of the
                   Association for Computational Linguistics},
  editor =        {Jurafsky, Dan and Chai, Joyce and Schluter, Natalie and
                   Tetreault, Joel},
  pages =         {8440--8451},
  publisher =     {Association for Computational Linguistics},
  title =         {Unsupervised Cross-lingual Representation Learning at
                   Scale},
  year =          {2020},
  doi =           {10.18653/v1/2020.acl-main.747},
  url =           {https://aclanthology.org/2020.acl-main.747/},
}
\appendix

\section*{Appendix}

\section{Baseline Prediction Analysis: \gemmaoneb{} vs.\ \gemmafourb{}}
\label{sec:baseline_analysis}

Table~\ref{tab:baseline_analysis} reports the baseline prediction statistics for \gemmaoneb{} and \gemmafourb{} on both evaluation datasets. The key pattern is that \gemmafourb{} systematically over-predicts the positive class: its predicted positive rate (65.7\% on ClaimBuster, 58.2\% on CLEF) far exceeds the true positive rate (26.5\% and 34.0\%), whereas \gemmaoneb{} is closer to the true distribution. This inflates cls-1 recall but suppresses cls-0 recall and overall weighted F1.

\begin{table}[H]
    \centering
    \small
    \begin{tabular}{lrrrr}
        \toprule
        & \multicolumn{2}{c}{ClaimBuster} & \multicolumn{2}{c}{CLEF 2024} \\
        \cmidrule(lr){2-3} \cmidrule(lr){4-5}
        Metric & \gemmaoneb & \gemmafourb & \gemmaoneb & \gemmafourb \\
        \midrule
        Gold positive rate (\%)   & 26.5 & 26.5 & 33.9 & 33.9 \\
        Predicted positive rate (\%) & 53.0 & 65.7 & 43.5 & 58.2 \\
        Mean score on gold-neg    & 0.38 & 0.48 & 0.30 & 0.42 \\
        Mean score on gold-pos    & 0.64 & 0.68 & 0.61 & 0.67 \\
        \midrule
        Cls-1 Precision           & 0.387 & 0.363 & 0.558 & 0.519 \\
        Cls-1 Recall              & 0.775 & 0.900 & 0.720 & 0.889 \\
        Cls-1 F1                  & 0.516 & 0.517 & 0.629 & 0.655 \\
        \midrule
        Cls-0 Precision           & 0.873 & 0.922 & 0.832 & 0.910 \\
        Cls-0 Recall              & 0.558 & 0.430 & 0.710 & 0.576 \\
        Cls-0 F1                  & 0.681 & 0.587 & 0.766 & 0.706 \\
        \bottomrule
    \end{tabular}
    \caption{Baseline prediction statistics for \gemmaoneb{} and \gemmafourb{}. Despite near-identical cls-1 F1 on ClaimBuster (0.516 vs.\ 0.517), \gemmafourb{} has far lower cls-0 recall (0.430 vs.\ 0.558) due to its higher predicted positive rate. The same positive-bias pattern holds on CLEF 2024.}
    \label{tab:baseline_analysis}
\end{table}

\section{Fine-Tuned Supervised Baseline: XLM-RoBERTa-Large}
\label{sec:xlm_roberta_ft}

We include a fine-tuned XLM-RoBERTa-Large model~\cite{conneau-etal-2020-unsupervised} as a supervised upper-bound baseline~\cite{amatya2026multilingualfactcheckingscalefinetuned}. Note that this is a multilingual model, since the startup provides multilingual fact-checking services. It is possible to achieve a better performance using a model optimized in a monolingual setting. The model was trained on a combined dataset of 84,312 examples drawn from ClaimBuster~\cite{hassan-2017-claimbuster}, CLEF 2024 CheckThat!~\cite{alam-2021-fighting-covid}, and in-house annotated claims (label 0: 43,979; label 1: 40,333). The model was fine-tuned using the training sets provided by these datasets to ensure that there is no leakage of test data in training. A linear classification head was added on top of the \texttt{[CLS]} token and trained with binary cross-entropy. To address class imbalance, a 5:1 positive-class weight was applied. Training used AdamW (lr $= 6{\times}10^{-6}$, weight decay $= 10^{-3}$), batch size 16, dropout 0.1, maximum sequence length 512, for up to 5 epochs with early stopping (patience 3) based on validation macro F1. The uncalibrated score $\hat{c}_i$ is the softmax probability of the positive class (label 1) if it exceeds 0.5, and the softmax probability of the negative class (label 0) otherwise, ensuring $\hat{c}_i \in [0,1]$ and compatibility with the residual correction in Eq.~\ref{eqn:ppi-decomp}. A detailed cost analysis comparing serving this model against frontier API models is provided in~\citealt{amatya2026multilingualfactcheckingscalefinetuned}.

\section{Model generation parameters used for the experiment}
\label{sec:model-parameters}

\begin{table}[hbt!]
    \centering
    \small
    \begin{tabular}{lrrrr}
        \hline
        Model & Temperature & Top-K & Top-P & Max Output Tokens \\
        \hline
        \gemmasmall  & 0.8 & 64 & 0.95 & default \\
        \gemmaoneb   & 1.0 & 64 & 0.95 & default \\
        \gemmafourb  & 1.0 & 64 & 0.95 & default \\
        \hline
        \gptmodel    & 1.0 & N/A & 1.0 & not set (max 128K) \\
        \claudeopus  & 1.0 & not set & not set & not set (max 128K) \\
        \hline
    \end{tabular}
    \caption{Generation parameters for all evaluated models. Gemma~3 models were served via Ollama with explicit sampling parameters. API-based models were queried with provider defaults.}
    \label{tab:generation_parameters}
\end{table}

\section{NN-PPI calibration evaluation at model temperature = 0.1}
\label{sec:nn-ppi-temp-0.1}

Table~\ref{tab:w_f1_with_low_temp} reports results when API-based and Gemma models are queried at $T{=}0.1$ instead of the default $T{=}1.0$. The overall pattern is consistent with the main results (Table~\ref{tab:ppi_weighted_scores}), confirming that \name{} gains \emph{are robust to this hyperparameter}.

\textbf{\gptmodel{}} is the only model where temperature has a negligible  effect. At $T{=}0.1$ the baseline weighted F1 improves from 0.843 to 0.853, driven by a Class~1 gain of +0.018 (0.685$\to$0.703) and a smaller Class~0 gain of +0.006 (0.901$\to$0.907). Lower temperature makes the model more deterministic, reducing sampling noise on genuinely check-worthy claims and improving CW recall. However, after \name{} calibration the gap largely closes: at $k{=}3$, \mknnppi{} Class~1 F1 is 0.653 vs.\ 0.659 at default temperature, a difference of only 0.006.

\textbf{\gemmaoneb{} and \gemmafourb{}} show no meaningful change across temperature settings. \gemmafourb{}'s baseline weighted F1 is identical (0.568) at both temperatures and both class F1 scores shift by at most 0.002. This confirms that the positive-class over-prediction bias of \gemmafourb{} is structural — encoded in the model weights — rather than a sampling artifact, and is therefore unaffected by temperature reduction. \gemmaoneb{} similarly shows negligible baseline movement ($\leq$0.003 across all metrics). \name{} calibration corrects both models' biases equally well at either temperature.

\begin{table*}[t!]
    \centering
    \scriptsize
    \begin{tabular}{llrrrrrrrrr}
        \toprule
        & & \multicolumn{3}{c}{Weighted F1} & \multicolumn{3}{c}{Class 0 F1} & \multicolumn{3}{c}{Class 1 F1} \\
        \cmidrule(lr){3-5} \cmidrule(lr){6-8} \cmidrule(lr){9-11}
        Model & $k$ & \mbaseline & \mknn & \mknnppi & \mbaseline & \mknn & \mknnppi & \mbaseline & \mknn & \mknnppi \\
        \midrule
        \multicolumn{11}{c}{\textit{ClaimBuster 2016}} \\
        \midrule
        \multirow[t]{3}{*}{\gptmodel} & 3 & \textbf{0.853} & 0.703 & 0.816 & \textbf{0.907} & 0.780 & 0.876 & \textbf{0.703} & 0.490 & 0.653 \\
         & 5 & \textbf{0.853} & 0.720 & 0.836 & \textbf{0.907} & 0.794 & 0.891 & \textbf{0.703} & 0.513 & 0.683 \\
         & 10 & \textbf{0.853} & 0.705 & \textbf{0.853} & \textbf{0.907} & 0.766 & 0.904 & 0.703 & 0.538 & \textbf{0.712} \\
        \addlinespace
        \midrule
        \multirow[t]{3}{*}{\gemmaoneb} & 3 & 0.635 & 0.704 & \textbf{0.731} & 0.677 & 0.780 & \textbf{0.803} & 0.518 & 0.491 & \textbf{0.531} \\
         & 5 & 0.635 & 0.719 & \textbf{0.739} & 0.677 & 0.794 & \textbf{0.810} & 0.518 & 0.512 & \textbf{0.543} \\
         & 10 & 0.635 & 0.706 & \textbf{0.742} & 0.677 & 0.766 & \textbf{0.808} & 0.518 & 0.539 & \textbf{0.561} \\
        \addlinespace
        \multirow[t]{3}{*}{\gemmafourb} & 3 & 0.568 & 0.703 & \textbf{0.733} & 0.587 & 0.780 & \textbf{0.806} & 0.515 & 0.490 & \textbf{0.531} \\
         & 5 & 0.568 & 0.720 & \textbf{0.746} & 0.587 & 0.795 & \textbf{0.817} & 0.515 & 0.513 & \textbf{0.550} \\
         & 10 & 0.568 & 0.706 & \textbf{0.763} & 0.587 & 0.766 & \textbf{0.829} & 0.515 & 0.539 & \textbf{0.580} \\
        \bottomrule
    \end{tabular}
    \caption{Weighted F1 for \mbaseline{} (few-shot), \mknn{}, and \mknnppi{} with \# of calibration samples $k=3,5,10$ and $T{=}0.1$. The observations for {\gptmodel}, {\gemmafourb} and {\gemmaoneb} at $T{=}0.1$ are consistent with default temperature results in Table~\ref{tab:ppi_weighted_scores}.}
    \label{tab:w_f1_with_low_temp}
    \vspace{-2em}
\end{table*}

\section{Ablation: Effect of labeled set size $|\mathcal{L}|$}
\label{sec:ablation-labeled-set-size}

\begin{figure}[H]
    \centering
    \includegraphics[width=1\linewidth]{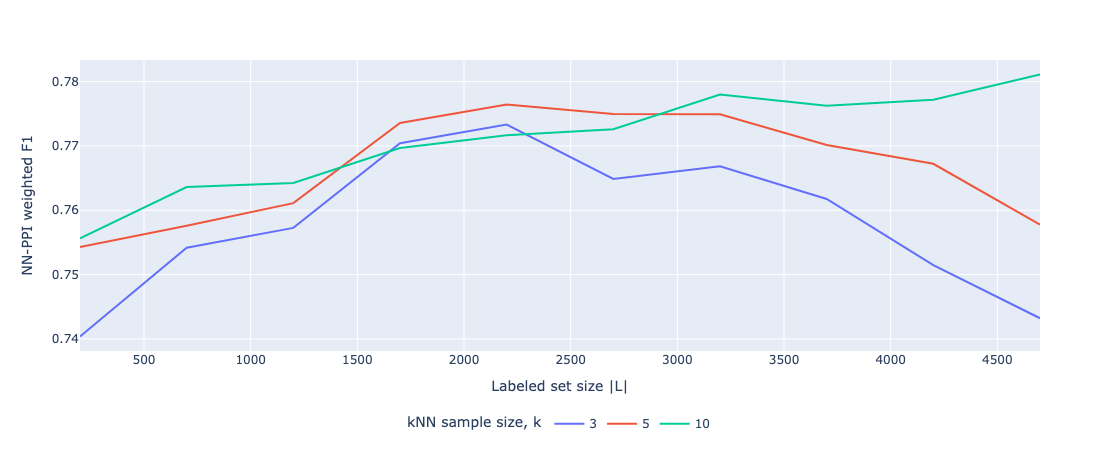}
    \caption{Weighted F1 of \name{} as a function of calibration set size $|\mathcal{L}|$ for $k \in \{3, 5, 10\}$. Performance stabilizes around $|\mathcal{L}|{=}1{,}500$--$2{,}000$ for $k{=}3$ and $k{=}5$, but continues to grow at $k{=}10$ beyond $|\mathcal{L}|{=}4{,}500$.}
    \label{fig:ablation-labeled-set-size}
\end{figure}

Figure~\ref{fig:ablation-labeled-set-size} shows weighted F1 as $|\mathcal{L}|$ is varied. For $k{=}3$ and $k{=}5$, performance peaks and stabilizes in the $|\mathcal{L}|{=}1{,}500$--$2{,}000$ range, motivating our choice of 1,314 (ClaimBuster) and 2,406 (CLEF 2024). For $k{=}10$, performance has not yet saturated at $|\mathcal{L}|{=}4{,}500$, suggesting that larger neighborhoods require a proportionally larger calibration pool to avoid residual noise from distributional mismatches; we leave this regime for future work.

\end{document}